\documentclass{article}
\usepackage{memory-idrr-arxiv-v1.0}

\usepackage{times}

\usepackage{amsmath}
\usepackage{amssymb}
\usepackage{amsopn}
\usepackage{latexsym}
\usepackage{subcaption}
\usepackage{CJKutf8}
\usepackage{graphicx}
\usepackage{multirow}
\usepackage{color}
\usepackage[usenames,dvipsnames]{xcolor}
\usepackage{tabularx}

\definecolor{zb_dred}{RGB}{102, 0, 0}
\definecolor{zb_red}{RGB}{255, 0, 0}
\definecolor{zb_lred}{RGB}{255, 102, 102} 

\title{Neural Discourse Relation Recognition with Semantic Memory}

\author{Biao Zhang$^{1,2}$, Deyi Xiong$^{2}$ and Jinsong Su$^{1}$\\
	Xiamen University, Xiamen, China 361005$^{1}$ \\
	Soochow University, Suzhou, China 215006$^{2}$ \\
	{\tt zb@stu.xmu.edu.cn, jssu@xmu.edu.cn} \\
	{\tt dyxiong@suda.edu.cn} \\
}

\begin{document}

\maketitle

\begin{abstract}


Humans comprehend the meanings and relations of discourses heavily relying on their semantic memory that encodes general knowledge about concepts and facts. Inspired by this, we propose a neural recognizer for implicit discourse relation analysis, which builds upon a semantic memory that stores knowledge in a distributed fashion. We refer to this recognizer as {\it SeMDER}. Starting from word embeddings of discourse arguments, SeMDER employs a {\it shallow encoder} to generate a distributed surface representation for a discourse. A {\it semantic encoder} with attention to the semantic memory matrix is further established over surface representations. It is able to  retrieve a deep semantic meaning representation for the discourse from the memory. Using the surface and semantic representations as input, SeMDER finally predicts implicit discourse relations via a {\it neural recognizer}. Experiments on the benchmark data set show that SeMDER benefits from the semantic memory and achieves substantial improvements of 2.56\% on average over current state-of-the-art baselines in terms of F1-score.

\end{abstract}

\section{Introduction}

Discourse relation recognition (DRR) that automatically identifies the logical relation of a coherent text is very important for discourse-level comprehension. It is relevant to a variety of nature language processing tasks such as summarization \cite{yoshida-EtAl:2014:EMNLP2014}, machine translation \cite{guzman-EtAl:2014:P14-1}, question answering \cite{jansen-surdeanu-clark:2014:P14-1} and information extraction \cite{cimiano2005ontology}. Although explicit DRR has recently achieved remarkable success \cite{miltsakaki2005experiments,pitler2008easily}, implicit DRR still remains a serious challenge due to the absence of discourse connectives.

However, even if discourse connectives are not provided, humans can still easily succeed in recognizing the relations of discourse arguments. One reason for this, according to cognitive psychology, would be that humans have a semantic memory in mind, which helps them comprehend word senses and further argument meanings via composition. After understanding what two arguments of a discourse convey, humans can easily interpret the  discourse relation of the two arguments. This semantic memory, as discussed by Tulving~\shortcite{tulving:episem}, refers to general knowledge including ``words and other verbal symbols, their meaning and referents, about relations among them, and about rules, formulas, and algorithms for manipulating them''. It can be retrieved to help disambiguation and comprehension whenever the barrier of cognition occurs. 

Consider the implicit discourse relation between the following two sentences:
\begin{quote}
(1)~{\it I was prepared to be in a very bad mood tonight.}

~\quad~{\it Now, I feel maybe there's a little bit of euphoria.}
\end{quote} 
It is difficult for conventional discourse relation recognizers to identify the relation between the two sentences as there is little significant surface information for use. However, if the recognizer obtains the knowledge of the antonymous relationship between the meaning of ``{\it bad mood}'' and that of ``{\it euphoria}'', it will be easy to infer the \textsc{comparison} relation between the two sentences. This semantic knowledge can be stored in an external memory for a discourse recognizer just like the semantic memory for humans.

 
\begin{figure}[t]
\centering
\includegraphics[scale=0.61]{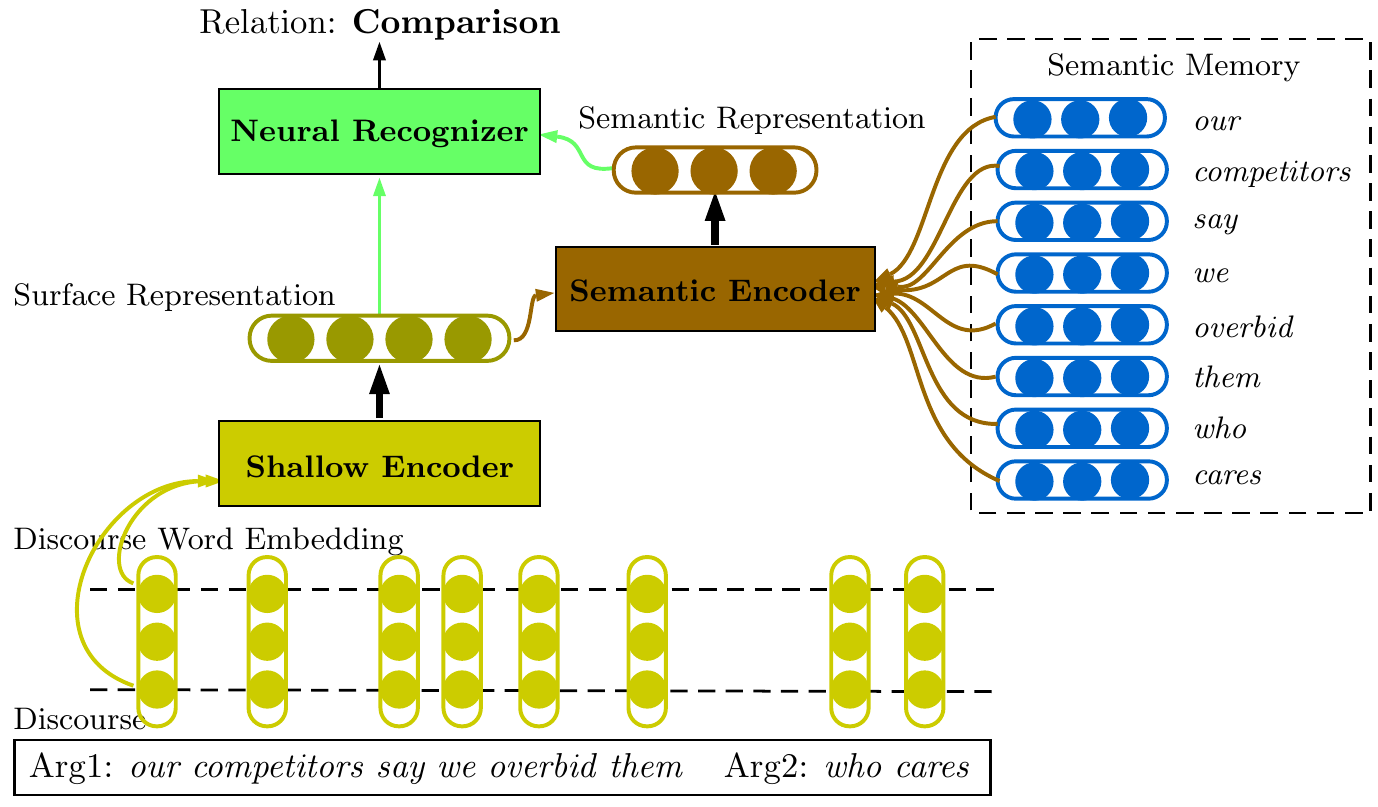}
\caption{\label{overal_model} Overall architecture for SeMDER model. We use the shallow and deep yellow color to indicate the surface and semantic representation respectively.}
\end{figure}


Inspired by the semantic memory in cognitive neuroscience \cite{yee2014cognitive} as well as memory network \cite{DBLP:journals/corr/WestonCB14,NIPS2015_5846,DBLP:journals/corr/KumarISBEPOGS15} and attentional mechanisms \cite{NIPS2014_5542,DBLP:journals/corr/BahdanauCB14,DBLP:conf/icml/XuBKCCSZB15}, we propose a neural network with semantic memory for implicit DRR, which refers to SeMDER. The philosophy of SeMDER includes: (1) the external semantic memory should be distributed as this allows easy computation; (2) the semantic memory should be easily accessed and retrieved; and (3) the retrieved content should be integrated into the comprehension of meanings of discourse arguments and their relations. In order to meet these requirements, we use a distributed matrix that encodes semantic knowledge of words as our external memory. The distributed memory is retrieved via an attentive reader. The retrieved distributed knowledge is incorporated into semantic representations of discourse arguments. Practically, we build a neural network that is composed of three essential components: a shallow encoder, a semantic encoder and a neural recognizer. The neural network is visualized in Figure \ref{overal_model}. In particular, 
\begin{itemize}
\item 
	{\bf Shallow encoder}: we feed word embeddings of discourse arguments into a {\it shallow encoder} \cite{biaozhang:2015:emnlp:drr} to obtain shallow representations of arguments. Due to their shallow property, we refer to them as surface representations (see Section \ref{scnn});
\item
	{\bf Semantic encoder}: we retrieve the semantic memory via an attention model. The retrieved content, together with surface representations, are incorporated into the semantic encoder to obtain deep semantic representations (see Section \ref{attention});
\item
	{\bf Neural recognizer}: both surface and semantic representations are feed into a {\it neural recognizer} to predict the corresponding discourse relations (see Section \ref{learning}).
\end{itemize}

Our contributions are twofold. First, we propose a neural network architecture for implicit DRR with an encoded semantic memory that enhances representations of arguments. To the best of our knowledge, we are the first to explore semantic memory for DRR via attentional mechanisms. Second, we conduct a series of experiments for English implicit DRR on the PDTB-style corpus to evaluate the effectiveness of our proposed neural network and semantic memory. Experiment results show that our network achieves substantial improvements against several strong baselines in term of F1 score. Extensive analysis on the attention further indicates that our model can recognize some important relation-relevant words, which we conjecture is the main reason for our success.

\section{Related Work}

The release of Penn Discourse Treebank (PDTB) \cite{prasad2008penn} opens the door to machine learning based implicit DRR. A variety of machine learning strategies have been presented previously, including feature engineering, connective predicting, data selection and discourse representation via neural networks.

Research on feature engineering exploits powerful and discriminative features for implicit DRR. In this respect, Pilter et al.~\shortcite{pitler-louis-nenkova:2009:ACLIJCNLP} investigate several linguistically informed features, such as polarity tags, verb classes, modality, context and lexical features. Lin et al.~\shortcite{lin2009recognizing} further consider contextual words, word pairs and parse trees for feature engineering. Later, several more powerful features have been developed: 
aggregated word pairs \cite{mckeown2013aggregated}, Brown clusters and coreference patterns \cite{rutherford-xue:2014:EACL}. With these features, Park and Cardie~\shortcite{park2012implicit} perform feature set optimization for better feature combination.

The major difference between explicit and implicit DRR is the presence of discourse connectives, the most salient features for DRR. Therefore, if we find a way to predict connectives for implicit discourses, we can transform implicit DRR into explicit DRR. Along this line, Zhou et al.~\shortcite{zhou2010predicting} use a language model to automatically insert discourse connectives, while Patterson and Kehler~\shortcite{patterson2013predicting} use a classifier to predict the presence or omission of a lexical connective. Different from this prediction strategy, Hong et al.~\shortcite{hong2012cross} leverage discourse connectives as a bridge between explicit and implicit relations and adopt an unsupervised cross-argument inference mechanism. 

Yet another strategy is data selection, where explicit discourse instances that are similar to the implicit ones are found and added to training corpus. Different data selection methods for implicit DRR can be classified into the following categories: instance typicality \cite{wang2012implicit}, multi-task learning \cite{lan2013leveraging}, domain adaptation \cite{braud-denis:2014:Coling,ji-zhang-eisenstein:2015:EMNLP}, semi-supervised learning \cite{Hernault:10:SemiSupervised,fisher-simmons:2015:ACL-IJCNLP} and explicit discourse connective classification \cite{rutherford-xue:2015:NAACL-HLT}. 

The third strategy is to learn representations of disourse arguments using neural networks for relation recognition, following remarkable success of neural networks in various natural language processing tasks. In this respect, Braud and Denis~\shortcite{ji-zhang-eisenstein:2015:EMNLP} investigates the usefulness of word representations. Specifically, two different neural network models have been developed for implicit DRR: recursive neural network for entity-augmented distributed semantics \cite{TACL536} and shallow convolutional neural network for discourse representation \cite{biaozhang:2015:emnlp:drr}. The former incorporates coreferent entity mentions into compositional distributed representations, while the latter develops a pure neural network model for discourse representations in implicit DRR. Normally, entities utilized in the former heavily depend on the availability and robustness of an upstream coreference system, and the latter only learns shallow representations for discourse arguments. Instead, our proposed model does not rely on any linguistic resources and incorporates a semantic memory to obtain deep semantic representations over shallow representations in \cite{biaozhang:2015:emnlp:drr}. Additionally, since the semantic memory is represented as a distributed matrix, our model is more robust and adaptable. 

The exploration of semantic memory for implicit DRR is inspired by recent developments in cognitive neuroscience. Yee et al.~\shortcite{yee2014cognitive} show how this memory is organized and retrieved in brain. In order to explore semantic memory in neural networks, we borrow ideas from recently introduced memory networks \cite{DBLP:journals/corr/WestonCB14,NIPS2015_5846,DBLP:journals/corr/KumarISBEPOGS15} to organize semantic memory as a distributed matrix and use an attention model to retrieve this distributed memory. The adaptation and utilization of semantic memory into implicit DRR, to the best of our knowledge, has never been investigated before.

\section{The SeMDER Model}

This section elaborates the proposed SeMDER model. We will first present the shallow encoder which converts a discourse into a distributed embedding. The semantic encoder, where the semantic memory is incorporated via an attention model is then described. After that, we explain how the neural recognizer classifies discourse relations. We also discuss the objective function and the procedure for parameter learning in this section.

\subsection{Shallow Encoder} \label{scnn}

To obtain surface representations for discourses, we employ a shallow convolutional neural network (SCNN) \cite{biaozhang:2015:emnlp:drr} as our shallow encoder. SCNN is specifically designed on the PDTB corpus, where implicit discourse relations are annotated between two neighboring arguments, namely {\it Arg1} and {\it Arg2} (see the example in Figure \ref{overal_model}). Given an argument which consists of $n$ words, SCNN represents each word as a $d$-dimensional dense, real-valued vector $x_i \in \mathbb{R}^{d_1}$ and concatenates them into a word embedding matrix:
\begin{equation}
\mathrm{X} = (x_{1}, x_{2}, \ldots, x_{n})
\end{equation}
where $\mathrm{X} \in \mathbb{R}^{d_1 \times n}$ forms the input layer of SCNN. All word vectors in vocabulary $V$ are stacked into a parameter matrix $L \in \mathbb{R}^{d_1 \times |V|}$ ($|V|$ is the vocabulary size), which will be tuned in the training phase.

To represent a discourse argument $c$, SCNN extracts major information inside $\mathrm{X}$ through three convolution operations {\it avg}, {\it min} and {\it max} defined as follows:
\begin{equation}
c_r^{avg} = \frac{1}{n} \sum_{i}^{n} X_{r,i}		
\end{equation}
\begin{equation}
c_r^{min} = \min \left( X_{r,1}, X_{r,2}, \ldots, X_{r,n} \right)
\end{equation}
\begin{equation}
c_r^{max} = \max \left(X_{r,1}, X_{r,2}, \ldots, X_{r,n} \right)
\end{equation}
where $r$ indicates the row of $\mathrm{X}$. The argument $c$ is thereby represented as the concatenation of these convolutional features:
\begin{equation}\label{argument}
c = \left[ c^{avg}; c^{max}; c^{min} \right]
\end{equation}
SCNN further obtains the representation $p \in \mathbb{R}^{6d_1}$ of a discourse by performing nonlinear transformations on the concatenation of two argument embeddings $c_{Arg1}$ and $c_{Arg2}$ generated in Eq. \ref{argument} as follows:
\begin{equation}
p = g(\left[c_{Arg1};c_{Arg2}\right]), ~~ g(x) = \frac{tanh(x)}{||tanh(x)||}
\end{equation}

Despite its simplicity, SCNN outperforms several feature-based systems. This is the reason that we choose it as our shallow encoder to obtain surface representations of discourses. However, the lack of deep knowledge in SCNN limits its further development. We therefore introduce a deep semantic encoder over the shallow encoder, which will be elaborated in the next section.

\subsection{Semantic Encoder} \label{attention}

Upon the surface representations, we further build a semantic encoder to incorporate a semantic memory to strengthen discourse comprehension. The semantic memory in SeMDER is represented as a distributed matrix $M \in \mathbb{R}^{m \times d_2}$ , where $d_2$ is the dimension of word embedding in the memory. Each row in the matrix indicates one word in discourse arguments (thus typically $m \leq n$). We assume that the semantic and syntactic attributes of words have already been encoded into this matrix. Therefore, incorporating this memory information into discourse representations will be beneficial for implicit DRR task.

\begin{figure}[t]
\centering
\includegraphics[scale=0.61]{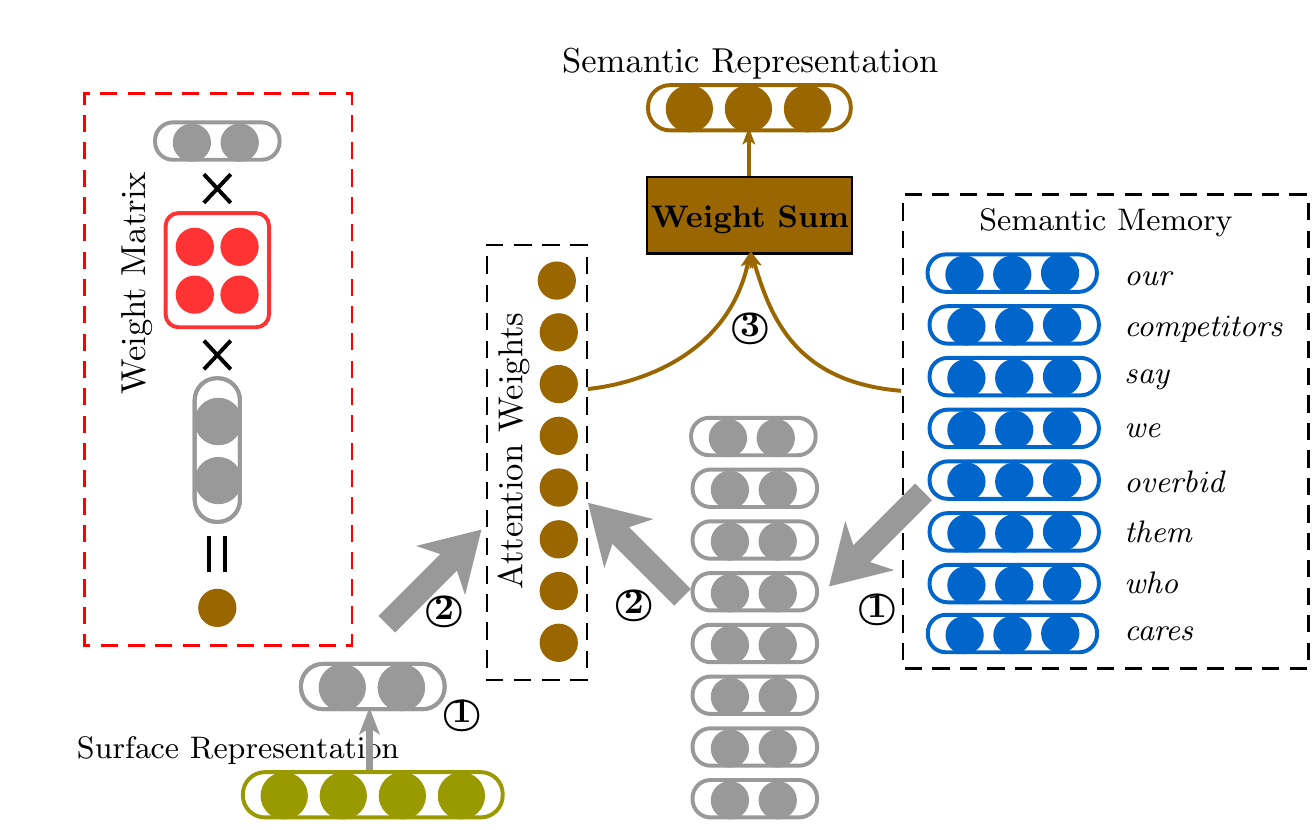}
\caption{\label{att_model} Illustration of the semantic encoder. We use gray color to indicate representations in the attention space. The dashed red box shows the bilinear-style computation for attention weights.}
\end{figure}

Figure \ref{att_model} gives an illustration of the procedure for incorporating the semantic memory. Specifically, given the surface representation $p$ for a discourse and the semantic memory matrix $M$, we stack an attention layer to project them onto the same space, which we call attention space. The projection is done as follows:
\begin{equation} \label{equ:p_a}
p_a = f(W_p p + b_a)
\end{equation}
\begin{equation} \label{equ:M_a}
M_a = f(W_m M^T + b_a)
\end{equation}
where the subscript $a$ denotes the attention space, $p_a$ and $M_a$ are the attentional representations for $p$ and $M$ respectively. $W_p \in \mathbb{R}^{d_a \times 6d_1}, W_m \in \mathbb{R}^{d_a \times d_2}$ are transformation matrices, $b_a \in \mathbb{R}^{d_a}$ is the bias term, $d_a$ is the dimensionality of the attention space, and $f(\cdot)$ is an element-wise activation function such as $tanh(\cdot)$, which is used throughout our model. The arrows marked by ``\textcircled{1}'' in Figure \ref{att_model} show this projection process.

Note that we differentiate the transformation matrix $W_p$ in Eq. \ref{equ:p_a} to the $W_m$ in Eq. \ref{equ:M_a}, since the surface representation and semantic memory are from different semantic spaces. However, we share the same bias term for them. This will force our model to learn to encode attention semantics into the transformation matrices, rather than the biases.

After obtaining the attentional representations for the discourse and semantic memory, we further estimate how useful each word memory cell $i$ in the semantic memory (i.e., the $i$th row in $M$) is to the corresponding discourse. This can be calculated by a match score:
\begin{equation}
s_i = g(p_a, M_{a, i}) \label{score}
\end{equation}
where $g(\cdot)$ is the scoring function. Since we are only interested in words occurred in the corresponding discourse, our attention schema is somewhat like a local attention. As discussed in \cite{DBLP:journals/corr/LuongPM15}, a {\it general} scoring function is much better for the local attention. Thus, we use a variant of the {\it general} function as our scoring function (see the red box in Figure \ref{att_model}):
\begin{equation}
g(p_a, M_{a, i}) = p_a W_s M_{a,i}
\end{equation}
where $W_s \in \mathbb{R}^{d_a \times d_a}$ is the bilinear scoring matrix, in which each element (see the red node in Figure \ref{att_model}) represents an interaction between the corresponding dimension of $p_a$ and $M_{a,i}$.

We further normalize the match score vector in Eq. \ref{score} to generate a probabilistic attention distribution over words in the semantic memory:
\begin{equation}
\alpha_i = \frac{exp(s_i)}{\sum_{j=1}^{m}exp(s_j)}
\end{equation}
Intuitively, the probability $\alpha_i$ (a.k.a attention weight) reflects the importance of the word $M_i$ in representing the whole discourse with respect to the final discourse relation recognition. Recall the above-mentioned example (1), if the importance of words ``{\it bad mood}'' and ``{\it euphoria}'' is recognized, there would be more chance that the final recognizer succeeds. 

Based on this attention distribution, we can compute the semantic representation for a discourse as a weighted sum of words in the semantic memory according to $\alpha$ (see the arrows marked by ``\textcircled{3}'' in Figure \ref{att_model}):
\begin{equation}\label{sum_words}
p_k = \sum_{j=1}^{m} \alpha_j M_{j}
\end{equation}
As shown in Eq. \ref{sum_words}, the semantic representation is directly retrieved from the semantic memory. It encodes semantic knowledge of words in discourse arguments that can help discourse relation recognition.

\subsection{Neural Recognizer} \label{learning}

Up to now, we have inferred both the surface and semantic representation for a discourse. To recognize the discourse relation, we further stack a Softmax layer upon these two representations:
\begin{equation}
y_p = h(W_{r,p} p + W_{r,k} p_k + b_r) 
\end{equation}
where $h(\cdot)$ is the softmax function, $W_{r,p} \in \mathbb{R}^{l \times 6d_1}, W_{r,k} \in \mathbb{R}^{l \times d_2}$ and $b_r \in \mathbb{R}^{l}$ are the parameter matrices and bias term respectively, and $l$ indicates the number of discourse relations.

\subsection{Objective Function and Parameter Learning}

Given a training corpus which contains $T$ instances $\{(x,y)\}_{t=1}^T$, we employ the following cross-entropy error to access how well the predicted relation $y_p$ represents the gold relation $y$,
\begin{equation}
E(y_p,y) = - \sum_{j}^{l} y_{j} \times \log\left(y_{p,j}\right)
\end{equation}
Therefore, the joint training objective of SeMDER is defined as follows:
\begin{equation}
J(\theta) = \frac{1}{T}\sum_{t=1}^{T} E(y_{p}^{(t)}, y^{(t)}) + R(\theta) \label{all_err}
\end{equation}
where $R(\theta)$ is the regularization term with respect to $\theta$. Towards the parameters $\theta$, we divide them into three different sets:
\begin{itemize}
\item
$\theta_L:$ word embedding matrix $L$;
\item
$\theta_R:$ discourse relation recognition parameters $W_{r,p}, W_{r,k}$ and $b_r$;
\item
$\theta_M:$ memory-related parameters $W_p, W_m, W_s$ and $b_a$;
\end{itemize}
All these parameters are regularized with corresponding weights\footnote{The bias terms $b$ is not regularized in practice.}:
\begin{equation}
R(\theta) = \frac{\lambda_L}{2}\|\theta_{L}\|^2 + \frac{\lambda_R}{2} \|\theta_{R}\|^2 + \frac{\lambda_M}{2} \|\theta_{M}\|^2
\end{equation}

Notice that although we can fine-tune the semantic memory in an end-to-end manner, we do not do that in our model. This is because we hope that the semantic and syntactic attributes encoded in the semantic memory can be preserved throughout our neural network.

We apply Limited-memory Broyden-Fletcher-Goldfarb-Shanno (L-BFGS) algorithm to optimize each parameter. In order to run the L-BFGS algorithm, we need to solve two problems: parameter initialization and partial gradient calculation.

In the phase of parameter initialization, $\theta_R$ and $\theta_M$ are randomly set according to a normal distribution ($\mu=0, \sigma=0.01$). For the word embedding $\theta_L$, we use the toolkit Word2Vec\footnote{https://code.google.com/p/word2vec/} to perform pretraining on a large-scale unlabeled data. This word embedding will be further fine-tuned in our SeMDER model to capture much more semantics related to discourse relations.

The partial gradient for parameter $\theta_j$ is computed as follows:
\begin{equation}
\frac{\partial J}{\partial \theta_j} = \frac{1}{T} \sum_{t=1}^T \frac{\partial E(y_{p}^{(t)}, y^{(t)})}{\partial \theta_j} +  \lambda_j \theta_j
\end{equation}
This gradient will be feed into the toolkit libLBFGS\footnote{http://www.chokkan.org/software/liblbfgs/} for parameter updating in our practical implementation.

\section{Experiments}

In this section, we conducted a series of experiments on English implicit DRR task. We begin with a brief review of the PDTB dataset. Then, we describe our experiment setup. Finally, we present experiment results and give an in-depth analysis on the attention.

\subsection{Dataset}

We used {\it PDTB 2.0} corpus\footnote{http://www.seas.upenn.edu/ pdtb/} \cite{prasad2008penn} (PDTB thereafter), which is the largest hand-annotated discourse corpus. Discourse relations are annotated in a predicate-argument view in PDTB, where each discourse connective is treated as a predicate that takes two text spans as its arguments. The relation tags in PDTB are arranged in a three-level hierarchy, where the top level consists of four major semantic \emph{classes}: \textsc{Temporal} (\textsc{Tem}), \textsc{Contingency} (\textsc{Con}), \textsc{Expansion} (\textsc{Exp}) and \textsc{Comparison} (\textsc{Com}). Because the top-level relations are general enough to be annotated with a high inter-annotator agreement and are common to most theories of discourse, in our experiments we only use this level of annotations.

\begin{table}[t]
\centering
\begin{tabular}{c|c|c|c}
\hline
\multirow{2}{*}{\bf Relation} & \multicolumn{3}{c}{\bf Positive/Negative Sentences} \\
\cline{2-4}
 & {\bf Train} & {\bf Dev} & {\bf Test} \\
\hline
\hline
\textsc{Com} & 1942/1942 & 197/986 & 152/894 \\
\textsc{Con} & 3342/3342 & 295/888 & 279/767 \\
\textsc{Exp} & 7004/7004 & 671/512 & 574/472 \\
\textsc{Tem} & 760/760 & 64/1119 & 85/96l \\
\hline		
\end{tabular}
\caption{\label{pdtb_data} Statistics of implicit discourse relations for the training (Train), development (Dev) and test (Test) sets in PDTB corpus.}
\end{table}

PDTB contains discourse annotations over 2,312 Wall Street Journal articles, and is organized in different sections. Following previous work \cite{pitler-louis-nenkova:2009:ACLIJCNLP,zhou2010predicting,lan2013leveraging,biaozhang:2015:emnlp:drr}, we used sections 2-20 as our training set, sections 21-22 as the test set. Sections 0-1 were used as the development set for hyperparameter optimization. We formulated the task as four separate one-against-all binary classification problems: each top level class vs. the other three discourse relation classes. We also balanced the training set by resampling training instances in each class until the number of positive and negative instances are equal. In contrast, all instances in the test and development set are kept in nature. The statistics of various data sets is listed in Table \ref{pdtb_data}.

\subsection{Setup}

We selected the {\it GoogleNews-vectors-negative300}\footnote{https://drive.google.com/file/d/0B7XkCwpI5KDYNlNUTTlSS\\ 21pQmM/edit?pref=2\&pli=1} as our external semantic memory. This data contains 300-dimensional vectors (thus, $d_2=300$) for 3 million words and phrases. It is trained on part of Google News dataset (about 100 billion words). The wide coverage and newswire domain of its training corpus as well as the syntactic property of word2vec models make this vector a good choice for the semantic memory.

We tokenized all datasets using {\it Stanford NLP Toolkit}\footnote{http://nlp.stanford.edu/software/corenlp.shtml}, and employed a large-scale unlabeled data\footnote{This data contains the training and development set for implicit DRR, as well as the English sentences in the FBIS corpus and the English sentences in Hansards part of LDC2004T07 corpus.} including 1.02M sentences (33.5M words) for word embedding $\theta_L$ initialization. We optimized the hyperparameters $d_1,\lambda_L,\lambda_R,\lambda_M$ according to previous work \cite{biaozhang:2015:emnlp:drr} and preliminary experiments on the development set. Finally, we set $d_1=128, \lambda_L=1e^{-5}, \lambda_R=\lambda_M=1e^{-4}$ for all the experiments. With respect to $d_a$, we tried three different settings $d_a=32,64,128$.

To validate the effectiveness of {\bf SeMDER} model, we compared against the following baseline methods:
\begin{itemize}
\item
{\bf SVM:} a support vector machine (SVM) classifier trained with the labeled data in the training set. We used the toolkit \emph{SVM-light}\footnote{http://svmlight.joachims.org/} to train the classifier in our experiments. 
\item
{\bf SCNN:} a shallow convolutional neural model proposed by Zhang et al.~\shortcite{biaozhang:2015:emnlp:drr}.
\end{itemize}
Features used in {\bf SVM} experiments are taken from the state-of-the-art implicit discourse relation recognition model, including \emph{Bag of Words}, \emph{Cross-Argument Word Pairs}, \emph{Polarity}, \emph{First-Last, First3}, \emph{Production Rules}, \emph{Dependency Rules} and \emph{Brown cluster pair} \cite{rutherford-xue:2014:EACL}. Additionally, in order to collect bag of words, production rules, dependency rules, and cross-argument word pairs, we used a frequency cutoff of 5 to remove rare features, following Lin et al.~\shortcite{lin2009recognizing}.

\subsection{Classification Results}

Because of the imbalance nature in our test set, we choose F1 score as our major evaluation metric. The performance of different models is presented in Table \ref{class_result}, which, overall, shows that {\bf SeMDER} outperforms the two baselines, achieving improvements in F1 score of 1.14\% on \textsc{Com}, 1.66\% on \textsc{Con}, 1.36\% on \textsc{Exp} and 5.62\% on \textsc{Tem} over the best baseline results.   We further observe that the improvements mainly result from high precision for \textsc{Com}, \textsc{Con} and \textsc{Tem}, while high recall for \textsc{Exp}. This is reasonable since the \textsc{Exp} relation owns the largest number of instances in our data.

\begin{table*}[t]
\centering
\begin{subtable}{.35\textwidth}
\centering
\begin{tabular}{c|c|c|c|c}
\hline
{\bf Model} & {$\mathrm{d_a}$} & {\bf P} & {\bf R} & {\bf F1} \\
\hline
\hline
{\bf SVM} & - & 22.79 & 64.47 & 33.68 \\ 

{\bf SCNN}	& - & 22.00 & 67.76 & 33.22 \\ 
\hline
\hline
\multirow{3}{*}{\bf SeMDER} 
	& 32 & 22.18 & 73.68 & 34.09 \\
	& 64 & 23.33 & 61.84 & 33.87 \\
	& 128 & 25.71 & 53.95 & {\bf 34.82} \\
\hline
\end{tabular}

\caption{\textsc{Com} vs Other}
\end{subtable}
\quad\quad~~~
\begin{subtable}{.35\textwidth}
\centering
\begin{tabular}{c|c|c|c|c}
\hline
{\bf Model} & {$\mathrm{d_a}$} & {\bf P} & {\bf R} & {\bf F1} \\
\hline
\hline
{\bf SVM} & - & 39.14 & 72.40 & 50.82 \\

{\bf SCNN}	& - & 39.80 & 75.29 & 52.04 \\ 
\hline
\hline
\multirow{3}{*}{\bf SeMDER}
    & 32  & 41.14 & 74.91 & 53.11 \\
    & 64  & 39.82 & 80.65 & 53.32 \\
    & 128 & 42.07 & 74.19 & {\bf 53.70} \\
\hline
\end{tabular}

\caption{\textsc{Con} vs Other}
\end{subtable}
\quad
\begin{subtable}{.35\textwidth}
\centering

\begin{tabular}{c|c|c|c|c}
\hline
{\bf Model} & {$\mathrm{d_a}$} &  {\bf P} & {\bf R} & {\bf F1} \\
\hline
\hline
{\bf SVM} & - & 65.89 & 58.89 & 62.19 \\

{\bf SCNN}	& - & 56.29 & 91.11 & 69.59 \\ 
\hline
\hline
\multirow{3}{*}{\bf SeMDER}
	& 32  & 54.80 & 99.48 & 70.67 \\
	& 64  & 54.79 & 99.65 & 70.70 \\
	& 128 & 54.98 & 100.0 & {\bf 70.95} \\
\hline
\end{tabular}

\caption{\textsc{Exp} vs Other}
\end{subtable}
\quad\quad~~~
\begin{subtable}{.35\textwidth}
\centering

\begin{tabular}{c|c|c|c|c}
\hline
{\bf Model} & {$\mathrm{d_a}$} & {\bf P} & {\bf R} & {\bf F1} \\
\hline
\hline
{\bf SVM} & - & 15.10 & 68.24 & 24.73 \\

{\bf SCNN}	& - & 20.22 & 62.35 & 30.54 \\ 
\hline
\hline
\multirow{3}{*}{\bf SeMDER} 
	& 32  & 21.79 & 60.00 & 31.97 \\
	& 64  & 23.01 & 61.18 & 33.44 \\
 	& 128 & 34.78 & 37.65 & {\bf 36.16} \\
\hline
\end{tabular}

\caption{\textsc{Tem} vs Other}
\end{subtable}

\caption{\label{class_result} Classification results of different models on implicit DRR. {\bf P}=Precision, {\bf R}=Recall, and {\bf F1}=F1 score. The best F1 scores are highlighted in bold.}
\end{table*}

\begin{table*}[t]
\centering
\begin{tabularx}{\textwidth}{c|X|l}
\hline
{\bf Relation} & {\bf Example} & {\bf Top Words}\\
\hline
\hline
\multirow{2}{*}{\textsc{Com}} & [people think of the steel business as an old and mundane smokestack business]$_{\emph{Arg1}}$, [they 're dead wrong]$_{\emph{Arg2}}$ & \multirow{2}{*}{\it wrong, people, dead, think, smokestack}\\
\hline
\multirow{2}{*}{\textsc{Con}} & [three minutes into the massage , the man curled up , began shaking and turned red]$_{\emph{Arg1}}$, [paramedics were called]$_{\emph{Arg2}}$ & \multirow{2}{*}{\it shaking, turned, paramedics, massage, curled} \\
\hline
\multirow{2}{*}{\textsc{Exp}} & [numerous injuries were reported]$_{\emph{Arg1}}$, [some buildings collapsed , gas and water lines ruptured and fires raged]$_{\emph{Arg2}}$ & \multirow{2}{*}{\it injuries, were, collapsed, raged, ruptured} \\
\hline
\multirow{2}{*}{\textsc{Tem}} & [warner sued sony and guber-peters late last week]$_{Arg1}$, [sony and guber-peters have countersued]$_{Arg2}$ & \multirow{2}{*}{\it have, countersued, late, week, last} \\ 
\hline
\end{tabularx}
\caption{\label{attention_analysis} Attention examples selected from the test set (we set $d_a=128$ for all relations). The top words are arranged in the order of attention weights.}
\end{table*}

As the neural baseline, {\bf SCNN} outperforms {\bf SVM} on \textsc{Con}, \textsc{Exp} and \textsc{Tem}, but fails on \textsc{Com}. The SeMDER with semantic memory, however, consistently surpasses {\bf SVM} and {\bf SCNN} in all discourse relations. This suggests that the incorporated semantic memory is helpful for recognizing correct discourse relations.  Additionally, for SeMDER, increasing the attention space dimensionality $d_a$ from 32 to 128 improves the performance in most cases.

Yet another interesting observation from Table \ref{class_result} is that the improvement of SeMDER over the two baselines for relation \textsc{Tem} is the biggest. The gain over {\bf SVM} is 11.4\% and 5.6\% over SCNN. This improvement is largely due to high precisions. As the number of instances in relation \textsc{Tem} is the smallest (see Table \ref{pdtb_data}), we argue that the traditional neural network models may suffer from overfitting in this case. However, our SeMDER enhanced with the semantic memory is capable of generalization that alleviates this overfitting issue.

\subsection{Attention Analysis}

We would like to know more about what role the semantic memory plays in our model, especially what the model learns from this semantic memory. Analyzing semantic representations is relatively meaningless. Therefore we turn to look into words with high attention weights for the answer. 

We present one example per discourse relation from the test set in Table \ref{attention_analysis}, where words assigned with the top-5 attention weights are listed separately. Consider the example for \textsc{Com}, our model retrieves the words ``{\it wrong, people, dead, think, smokestack}'', which roughly reflect the discourse meaning that {\it people think smokestack, dead wrong}. Obviously, these words are crucial for discourse comprehension. These examples display that SeMDER prefers to retrieve from the semantic memory relation-relevant words that strongly indicate the corresponding relations, which we think is the main reason for the success of SeMDER.


\section{Conclusion and Future Work}

In this paper, we have presented a neural discourse relation recognizer with a distributed semantic memory for implicit DRR. The semantic memory encodes semantic knowledge of words in discourse arguments and helps disambiguation and comprehension. We employ an attention model to retrieve discourse relation-relevant information into semantic representations of discourses, which, to some extend, simulates the cognition process of humans. Experiment results show that our model outperforms several strong baselines, and further analysis reveals that our model can indeed detect some relation-relevant words.

In the future, we would like to exploit different types of semantic memory, e.g., a distributed memory on ontology concepts and relations. We also want to explore different attention architectures, e.g. the {\it concat} and {\it dot} in \cite{DBLP:journals/corr/LuongPM15}. Furthermore, we are interested in adapting our model to other similar classification tasks, such as sentiment classification, movie review
classification and nature language inference.

\newpage

{\small
\bibliographystyle{named}
\bibliography{memory-idrr-arxiv-v1.0}
}
\end{document}